\documentclass[10pt,twocolumn,letterpaper]{article}

 \usepackage[pagenumbers]{iccv} % To force page numbers, e.g. for an arXiv version

\usepackage{amsmath} 
\usepackage{quiver}
\usepackage{graphicx}

\usepackage{colortbl} % for row coloring
\definecolor{iccvblue}{rgb}{0.21,0.49,0.74}
\usepackage[pagebackref,breaklinks,colorlinks,allcolors=iccvblue]{hyperref}

\def\paperID{*****} % *** Enter the Paper ID here
\def\confName{ICCV}
\def\confYear{2025}

\title{A Survey on Training-free Open-Vocabulary Semantic Segmentation}

\author{
Naomi Kombol \quad \quad \quad
Ivan Martinović \quad \quad \quad
Siniša Šegvić \\
Sveučilište u Zagrebu, Fakultet elektrotehnike i računarstva \\
{\tt\small naomi.kombol@fer.hr, ivan.martinovic@fer.hr, sinisa.segvic@fer.hr}
}

\begin{document}
\maketitle
\begin{abstract}

Semantic segmentation is one of the most fundamental tasks in image understanding with a long history of research, and subsequently a myriad of different approaches. Traditional methods strive to train models up from scratch, requiring vast amounts of computational resources and training data. In the advent of moving to open-vocabulary semantic segmentation, which asks models to classify beyond learned categories, large quantities of finely annotated data would be prohibitively expensive. Researchers have instead turned to training-free methods where they leverage existing models made for tasks where data is more easily acquired. Specifically, this survey will cover the history, nuance, idea development and the state-of-the-art in training-free open-vocabulary semantic segmentation that leverages existing multi-modal classification models. We will first give a preliminary on the task definition followed by an overview of popular model archetypes and then spotlight over 30 approaches split into broader research branches: purely CLIP-based, those leveraging auxiliary visual foundation models and ones relying on generative methods. Subsequently, we will discuss the limitations and potential problems of current research, as well as provide some underexplored ideas for future study. We believe this survey will serve as a good onboarding read to new researchers and spark increased interest in the area.

\vspace{0.5em}

\end{abstract}

\section{Introduction}
\label{sec:intro}

Image segmentation stands as one of the most prolific areas of research in computer vision and is a core component of comprehensive image understanding. It enjoys widespread use in medical imaging \citep{MedSAM}, autonomous driving \citep{nuScenes}, agriculture \citep{agriCowCount, agriDiesaseSurvey}, industry \citep{indFruitDetector, indMetalCorrosion}, and many more areas.

Segmenting pixels into semantically coherent categories has a long, at first non-machine-learning based, history of study. Initial attempts relied on thresholding \cite{Otzu} and edge detection \cite{Canny}, then moved to region-based methods \cite{Watershed}. Markov and conditional random fields rose to prominence for a time \cite{MRF_CRF}, as well as graph-based methods \cite{Graph_based}, before deep learning took point. Fully convolutional networks \cite{FCN}, encoder-decoder based approaches \cite{UNet}, exploiting dilated convolutions \cite{DeepLab} and then expanding the Faster R-CNN architecture for instance segmentation \cite{Mask_RCNN} followed. The most modern methods rely on transformers \cite{SETR, MaskFormer, Mask2Former}, and constitute the edge of exploration.

Beyond the stifling constraints of finite class sets, work naturally advanced to tasks where segmentation was performed on arbitrary category sets \cite{zero-shot, OV-M-S, GroupVit}. The open-vocabulary setting was made possible by vision-language models (VLM-s), capable of embedding textual class representations into a shared semantic space with visual information: enabling direct comparison and subsequent grouping.

Training such models usually demands heavy computational resources and vast, expensive amounts of data in exchange for the impressive results. An alternative course of research aims to downstream pretrained VLM-s, such as CLIP \cite{CLIP}, to segmentation without additional training. By cleverly modifying architecture \citep{MaskCLIP, ClipSurgery, SCLIP} and manipulating the knowledge obtained during contrastive training for robust classification, CLIP can be adapted to dense prediction.

To the best of our knowledge, this survey offers an exhaustive overview of the state of training-free open-vocabulary semantic segmentation. Methods will be divided into broad research directions for easier comprehension and delved into individually to study their nuances and innovations. Categories are organized chronologically to emphasize the chains of ideas that make up the most advanced approaches.

\section{Preliminaries}
\label{sec:relatedwork}
This section will offer a brief summary of vision-language models (VLM-s), visual foundation models (VFM-s) and give a clearer description of the examined problem to establish common terms and ground.

%-------------------------------------------------------------------------

\subsection{Vision-language Models}
The idea of jointly understanding visual stimuli and natural language made great strides in advancing the zero-shot and subsequently open-vocabulary capabilities of models \cite{ALIGN, CLIP, SIGLIP}. One of the most popular examples is CLIP \cite{CLIP}: contrastively pretrained on millions of image-caption pairs to learn proper alignment and association between corresponding true matches. The extensive training and so derived knowledge facilitate great generalization capabilities and impressive zero-shot performance on classification tasks.

\subsection{Visual Foundation Models}
Real life objects inherently possess spatial consistency, although such knowledge is not necessarily explicitly taught to models. To counteract that deficiency, a new archetype emerged. Visual Foundation Models (VMF-s) pre-train on large-scale datasets to produce structurally aware and consistent features that are suitable for down-streaming to various visual tasks: from object detection \cite{TokenCut} to semantic segmentation \cite{DINO, DINOv2, SAM}.

Self-supervised training is one tool to ensure that without costly annotations. A model learns meaningful information from raw images by making predictions that do not require explicit labeling. Instead, the inherent order and structure of the data act as training signals. The DINO line \cite{DINO, DINOv2} champions this approach by constructing space-aware embeddings.

On the other hand, SAM \cite{SAM} trains in a supervised manner and generates class-agnostic masks of objects present in images.

\subsection{Problem Definition: Open-vocabulary Semantic Segmentation}

Semantic segmentation is the natural extension of classification; instead of only knowing what is in an image, we would like to know precise locations. It is a fundamental computer vision task that dictates classifying each pixel of an image into a semantic category.
Formally: given the input image \( I \in \mathbb{R}^{H \times W \times C} \), where \( H \) and \( W \) represent the height and width, and \( C \) is the number of channels (usually "RGB"), we aim to obtain the label map \( L \in \mathbb{R}^{H \times W} \) where each element \( L_{i,j} \) denotes the class of the pixel at position \( (i,j) \).

The open-vocabulary prerequisite demands a model be able to segment arbitrary visual concepts based on equally arbitrary textual class representations; irrespective of if they've been seen during training \cite{zero-shot, GroupVit}. Language-driven segmentation approaches are proving to be a region of growing interest.

%-------------------------------------------------------------------------

\section{Training-free Open-Vocabulary Semantic Segmentation}
\label{sec:tfovss} 

In order to leverage the maximum potential of vision-language pre-trained classification models, a novel idea of adapting existing architecture to dense prediction was born. By minimally altering the inference pipeline, this approach incurs no additional training cost, yet allows for full image segmentation.

\begin{figure}[h]  % Use [h] to keep it in the column
    \centering
    
\tikzset{every picture/.style={line width=0.75pt}} %set default line width to 0.75pt        

\begin{tikzpicture}[x=0.75pt,y=0.75pt,yscale=-1,xscale=1]
%uncomment if require: \path (0,235); %set diagram left start at 0, and has height of 235

%Rounded Rect [id:dp8664242432572626] 
\draw  [color={rgb, 255:red, 233; green, 113; blue, 50 }  ,draw opacity=1 ][fill={rgb, 255:red, 200; green, 72; blue, 12 }  ,fill opacity=1 ] (205.2,83.13) .. controls (205.2,79.34) and (208.27,76.27) .. (212.06,76.27) -- (274.27,76.27) .. controls (278.06,76.27) and (281.13,79.34) .. (281.13,83.13) -- (281.13,103.69) .. controls (281.13,107.48) and (278.06,110.55) .. (274.27,110.55) -- (212.06,110.55) .. controls (208.27,110.55) and (205.2,107.48) .. (205.2,103.69) -- cycle ;
%Rounded Rect [id:dp5908819016085214] 
\draw  [color={rgb, 255:red, 233; green, 113; blue, 50 }  ,draw opacity=1 ][fill={rgb, 255:red, 233; green, 113; blue, 50 }  ,fill opacity=1 ] (205.84,136.43) .. controls (205.84,134.7) and (207.24,133.3) .. (208.97,133.3) -- (278.79,133.3) .. controls (280.52,133.3) and (281.93,134.7) .. (281.93,136.43) -- (281.93,145.83) .. controls (281.93,147.56) and (280.52,148.96) .. (278.79,148.96) -- (208.97,148.96) .. controls (207.24,148.96) and (205.84,147.56) .. (205.84,145.83) -- cycle ;
%Rounded Rect [id:dp2879899738192895] 
\draw  [color={rgb, 255:red, 233; green, 113; blue, 50 }  ,draw opacity=1 ][fill={rgb, 255:red, 233; green, 113; blue, 50 }  ,fill opacity=0.76 ] (293.91,51.33) .. controls (293.91,49.6) and (295.32,48.2) .. (297.05,48.2) -- (392.63,48.2) .. controls (394.36,48.2) and (395.76,49.6) .. (395.76,51.33) -- (395.76,60.73) .. controls (395.76,62.46) and (394.36,63.86) .. (392.63,63.86) -- (297.05,63.86) .. controls (295.32,63.86) and (293.91,62.46) .. (293.91,60.73) -- cycle ;
%Rounded Rect [id:dp31766289502129497] 
\draw  [color={rgb, 255:red, 233; green, 113; blue, 50 }  ,draw opacity=1 ][fill={rgb, 255:red, 233; green, 113; blue, 50 }  ,fill opacity=0.76 ] (293.01,136.93) .. controls (293.01,135.2) and (294.42,133.8) .. (296.15,133.8) -- (392.13,133.8) .. controls (393.86,133.8) and (395.26,135.2) .. (395.26,136.93) -- (395.26,146.33) .. controls (395.26,148.06) and (393.86,149.46) .. (392.13,149.46) -- (296.15,149.46) .. controls (294.42,149.46) and (293.01,148.06) .. (293.01,146.33) -- cycle ;
%Rounded Rect [id:dp754415500117372] 
\draw  [color={rgb, 255:red, 233; green, 113; blue, 50 }  ,draw opacity=1 ][fill={rgb, 255:red, 233; green, 113; blue, 50 }  ,fill opacity=0.76 ] (293.51,189.17) .. controls (293.51,187.44) and (294.92,186.04) .. (296.65,186.04) -- (393.56,186.04) .. controls (395.29,186.04) and (396.69,187.44) .. (396.69,189.17) -- (396.69,198.57) .. controls (396.69,200.3) and (395.29,201.7) .. (393.56,201.7) -- (296.65,201.7) .. controls (294.92,201.7) and (293.51,200.3) .. (293.51,198.57) -- cycle ;
%Straight Lines [id:da5725192884169401] 
\draw    (281.66,141.13) -- (287.09,141.13) -- (293.12,141.13) ;
%Straight Lines [id:da7419873953625526] 
\draw    (287.25,193.72) -- (287.25,56.06) ;
%Straight Lines [id:da6881011343355563] 
\draw    (287.25,56.03) -- (293.91,56.03) ;
%Straight Lines [id:da7375150501782215] 
\draw    (287.25,193.72) -- (293.12,193.72) ;
%Rounded Rect [id:dp4220648905781683] 
\draw  [color={rgb, 255:red, 233; green, 113; blue, 50 }  ,draw opacity=1 ][fill={rgb, 255:red, 233; green, 113; blue, 50 }  ,fill opacity=0.5 ] (409.05,72.9) .. controls (409.05,71.17) and (410.45,69.77) .. (412.18,69.77) -- (517.3,69.77) .. controls (519.03,69.77) and (520.43,71.17) .. (520.43,72.9) -- (520.43,82.3) .. controls (520.43,84.03) and (519.03,85.43) .. (517.3,85.43) -- (412.18,85.43) .. controls (410.45,85.43) and (409.05,84.03) .. (409.05,82.3) -- cycle ;
%Rounded Rect [id:dp8845110167896888] 
\draw  [color={rgb, 255:red, 233; green, 113; blue, 50 }  ,draw opacity=1 ][fill={rgb, 255:red, 233; green, 113; blue, 50 }  ,fill opacity=0.5 ] (408.77,94.18) .. controls (408.77,92.45) and (410.17,91.04) .. (411.9,91.04) -- (518.15,91.04) .. controls (519.88,91.04) and (521.28,92.45) .. (521.28,94.18) -- (521.28,103.57) .. controls (521.28,105.3) and (519.88,106.7) .. (518.15,106.7) -- (411.9,106.7) .. controls (410.17,106.7) and (408.77,105.3) .. (408.77,103.57) -- cycle ;
%Rounded Rect [id:dp45647658443029715] 
\draw  [color={rgb, 255:red, 233; green, 113; blue, 50 }  ,draw opacity=1 ][fill={rgb, 255:red, 233; green, 113; blue, 50 }  ,fill opacity=0.5 ] (409.05,115.6) .. controls (409.05,113.87) and (410.45,112.47) .. (412.18,112.47) -- (518.15,112.47) .. controls (519.88,112.47) and (521.28,113.87) .. (521.28,115.6) -- (521.28,124.99) .. controls (521.28,126.72) and (519.88,128.13) .. (518.15,128.13) -- (412.18,128.13) .. controls (410.45,128.13) and (409.05,126.72) .. (409.05,124.99) -- cycle ;
%Rounded Rect [id:dp9459945351438788] 
\draw  [color={rgb, 255:red, 233; green, 113; blue, 50 }  ,draw opacity=1 ][fill={rgb, 255:red, 233; green, 113; blue, 50 }  ,fill opacity=0.5 ] (408.49,51.33) .. controls (408.49,49.6) and (409.89,48.2) .. (411.62,48.2) -- (517.3,48.2) .. controls (519.03,48.2) and (520.43,49.6) .. (520.43,51.33) -- (520.43,60.73) .. controls (520.43,62.46) and (519.03,63.86) .. (517.3,63.86) -- (411.62,63.86) .. controls (409.89,63.86) and (408.49,62.46) .. (408.49,60.73) -- cycle ;
%Rounded Rect [id:dp41544720398792734] 
\draw  [color={rgb, 255:red, 233; green, 113; blue, 50 }  ,draw opacity=1 ][fill={rgb, 255:red, 233; green, 113; blue, 50 }  ,fill opacity=0.5 ] (408.2,136.72) .. controls (408.2,134.99) and (409.61,133.59) .. (411.34,133.59) -- (518.15,133.59) .. controls (519.88,133.59) and (521.28,134.99) .. (521.28,136.72) -- (521.28,146.12) .. controls (521.28,147.85) and (519.88,149.25) .. (518.15,149.25) -- (411.34,149.25) .. controls (409.61,149.25) and (408.2,147.85) .. (408.2,146.12) -- cycle ;
%Rounded Rect [id:dp8697699116143247] 
\draw  [color={rgb, 255:red, 233; green, 113; blue, 50 }  ,draw opacity=1 ][fill={rgb, 255:red, 233; green, 113; blue, 50 }  ,fill opacity=0.5 ] (409.05,158) .. controls (409.05,156.27) and (410.45,154.87) .. (412.18,154.87) -- (518.57,154.87) .. controls (520.3,154.87) and (521.7,156.27) .. (521.7,158) -- (521.7,167.4) .. controls (521.7,169.12) and (520.3,170.53) .. (518.57,170.53) -- (412.18,170.53) .. controls (410.45,170.53) and (409.05,169.12) .. (409.05,167.4) -- cycle ;
%Straight Lines [id:da7997764010909871] 
\draw    (396.27,141.19) -- (401.22,141.16) -- (401.06,162.17) -- (409.06,162.17) ;
%Straight Lines [id:da2815149143315583] 
\draw    (399.95,119.26) -- (399.95,56.06) ;
%Straight Lines [id:da693028593617935] 
\draw    (395.95,56.03) -- (408.2,56.03) ;
%Straight Lines [id:da412348462808632] 
\draw    (399.95,77.3) -- (409.8,77.3) ;
%Straight Lines [id:da563972840237013] 
\draw    (400.16,97.72) -- (409.22,97.72) ;
%Straight Lines [id:da9368778788309493] 
\draw    (399.95,119.26) -- (409,119.26) ;
%Straight Lines [id:da6058247342685346] 
\draw    (401.22,141.16) -- (408.58,141.13) ;
%Straight Lines [id:da5551061166937423] 
\draw    (243.3,133) -- (243.3,110.58) ;

% Text Node
\draw (214.73,82.8) node [anchor=north west][inner sep=0.75pt]  [font=\fontsize{0.43em}{0.52em}\selectfont] [align=left] { \ \ \ \ \ \ \ Training Free\\ \ \ \ \ Open-Vocabulary\\Semantic Segmentation};
% Text Node
\draw (213.17,138.39) node [anchor=north west][inner sep=0.75pt]  [font=\fontsize{0.43em}{0.52em}\selectfont] [align=left] {CLIP-Based Approaches};
% Text Node
\draw (297.89,53.3) node [anchor=north west][inner sep=0.75pt]  [font=\fontsize{0.43em}{0.52em}\selectfont] [align=left] {Purely CLIP-Based (\ref{subsec:clip_based})};
% Text Node
\draw (296.78,139.25) node [anchor=north west][inner sep=0.75pt]  [font=\fontsize{0.43em}{0.52em}\selectfont] [align=left] {Use VFM-s alongside CLIP (\ref{subsec:vfms})
};
% Text Node
\draw (294.47,190.49) node [anchor=north west][inner sep=0.75pt]  [font=\fontsize{0.43em}{0.52em}\selectfont] [align=left] {Generative Methods Alongside CLIP (\ref{subsec:gen_methods})};
% Text Node
\draw (411.97,74.87) node [anchor=north west][inner sep=0.75pt]  [font=\fontsize{0.43em}{0.52em}\selectfont] [align=left] {Leverage Intermediate Layers (\ref{subsubsec:clip_intermed})};
% Text Node
\draw (410.74,96.18) node [anchor=north west][inner sep=0.75pt]  [font=\fontsize{0.43em}{0.52em}\selectfont] [align=left] {Leverage Non-ML Obtained Masks (\ref{subsubsec:clip_mask})};
% Text Node
\draw (410.7,117.12) node [anchor=north west][inner sep=0.75pt]  [font=\fontsize{0.43em}{0.52em}\selectfont] [align=left] {Others (\ref{subsubsec:clip_misc})};
% Text Node
\draw (412.04,53.3) node [anchor=north west][inner sep=0.75pt]  [font=\fontsize{0.43em}{0.52em}\selectfont] [align=left] {Refine Inter-token Mixing (\ref{subsubsec:clip_attn})
};
% Text Node
\draw (411.06,138.69) node [anchor=north west][inner sep=0.75pt]  [font=\fontsize{0.43em}{0.52em}\selectfont] [align=left] {Refine Inter-token Mixing with VFMs (\ref{subsubsec:vfm_attn})};
% Text Node
\draw (410.8,159.96) node [anchor=north west][inner sep=0.75pt]  [font=\fontsize{0.43em}{0.52em}\selectfont] [align=left] {Leverage VFMs for Maskpooling (\ref{subsubsec:vfm_maskpooling})};

\end{tikzpicture}

\tikzset{every picture/.style={line width=0.75pt}} %set default line width to 0.75pt        
    \caption{Overview of Training-free Open-Vocabulary Semantic Segmentation}
    \label{fig:ovss_diagram}
\end{figure}
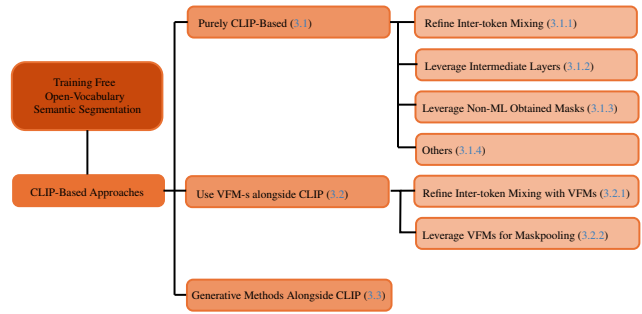

This section will comprehensively cover current strains of research in training-free open-vocabulary semantic segmentation (OVSS) with care given to present the evolution of ideas over time as well. As seen in \cref{fig:ovss_diagram}, we will first go over models based purely on CLIP \cite{CLIP} and how knowledge instilled in the original weights can be used to refine inter-token mixing, how information from intermediate layers can aid decision-making, how grouping CLIP embeddings can be utilized and a few other methods. Afterward, we will cover the use of auxiliary VFM-s in tandem with CLIP to further refine attention and showcase some more advanced processes that rely on mask-pooling. The last branch of study utilizes generative models, mostly to obtain either visual or textual prototypes, or to scrutinize language-driven generation itself to obtain relationships between image and text.

\subsection{Transformer CLIP Based Approaches} \label{subsec:clip_based}
\begin{figure}[htb]
    \centering
    \includegraphics[width=0.26\columnwidth]{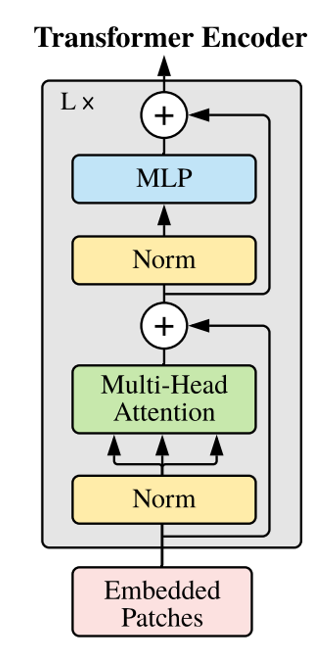}
    \caption{Standard ViT encoder block that first normalizes inputs and then performs inter-token mixing through multi-head attention. The remixed tokens are residually connected to the inputs, the sum normalized, projected through a Multi-Layer Perceptron (MLP), and again residually connected. Image is from \cite{vit}}
    \label{fig:encoder}
\end{figure}

Vision Transformers \citep{vit} (ViT-s) are a subtype of transformers \cite{Transformer} adapted to parsing visual information. An image is first divided into square patches, usually of size 16x16, which are subsequently encoded as tokens and enriched with positional embeddings to preserve spatial relationships. Such tokens are passed through multiple standard encoder blocks (as seen in \cref{fig:encoder}) to allow for information refinement and exchange. Depending on the task, the embeddings at the end of the encoder are taken as is or further processed.

The standard CLIP \cite{CLIP} model, made for classification tasks, ends its visual encoder with a typical block (\cref{fig:encoder}) where attention calculation follows \cref{eq:attn_normal}.

\begin{equation}
    \text{\textbf{X'}} = \operatorname{softmax} \left( \mathbf{X} \mathbf{W}_q \mathbf{W}_k^T \mathbf{X}^T / \tau \right)  \mathbf{W}_v \mathbf{X}
    \label{eq:attn_normal}
\end{equation}

First, the input embeddings are projected into the query and key spaces for similarity comparison. Based upon how alike they are, value-projected inputs are inter-combined to form new, richer embeddings. The first token in the \( X' \) sequence, i.e. the classification token [CLS], is entrusted with encoding global information about the image that will be projected into the same semantic space as text embeddings. Classification is done by comparing how alike the [CLS] token is with the embeddings of textual class representations. The equation is simplified to show only one head for convenience. Matrices $W_q$, $W_k$ and $W_v$ represent weights for the query, key and value projections respectively, $X$ denotes the inputs to the attention layer and $\tau$ is the temperature. Since the final inter-token mixing enriches the semantics of the [CLS] token based not only on itself but the tokens corresponding to the original patches, the information they contain has to be semantically relevant as well: it stands to reason they could be exploited for dense prediction.

The primeval work of exploring CLIP for such a task was MaskCLIP \cite{MaskCLIP} which first adapted the basic visual transformer architecture of CLIP for arbitrary semantic segmentation. By extracting all the tokens \textit{except} the [CLS] token from the last attention layer of the transformer, normalizing and passing them through the last linear projection to the shared visual-language space, MaskCLIP demonstrates how to obtain per-patch classifiable embeddings. They are then compared to the textual class embeddings for segmentation. 
We refer to this baseline as CLIPBase. \cref{tab:divided_tfovss} shows its promise, with the numbers being noticeably better than random, and in \cref{fig:voc21s} we can see that although localization is not good, the requisite categories are mostly present.

MaskCLIP suggests that the problem is in the inter-token mixing step of the final encoder layer i.e. attention. Instead of using projections to query and key space when calculating similarity, they force the attention weights to an identity matrix (\cref{eq:attn_maskclip}), stressing the importance of locality and similarity in the same semantic space, which yields considerable improvement (\cref{tab:divided_tfovss}). \cref{fig:voc21s} showcases tentatively observable outlines of the pictured objects and how the global information is not as overpowering as with the CLIPBase approach.

\begin{equation}
    \text{\textbf{Attn}}_{\text{MaskCLIP}} = \mathbf{I} \cdot \mathbf{W}_v \mathbf{X}
    \label{eq:attn_maskclip}
\end{equation}

\subsubsection{Refining Attention} \label{subsubsec:clip_attn}

\begin{figure}[htb]
    \centering
    \includegraphics[width=0.25\columnwidth]{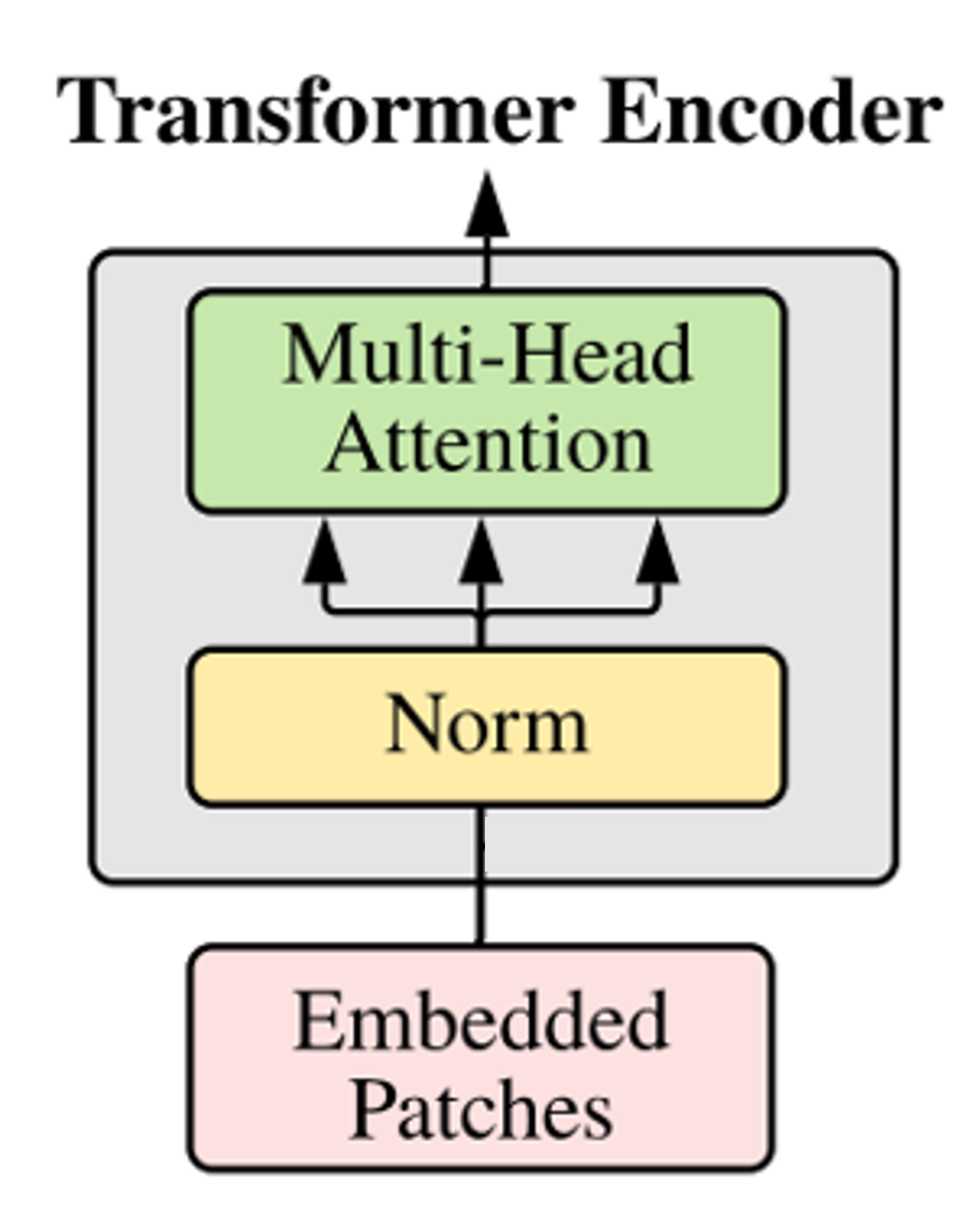}
    \caption{Modified ViT encoder block with removed MLP and residual connection. Image is from \cite{vit} and modified.}
    \label{fig:encoder_modified}
\end{figure}

The area of research with the most work has been improving performance through altering attention in varying places and maximizing the utilization of preexisting CLIP models \cite{ClipSurgery, SCLIP, GEM, ClearCLIP, NACLIP, ITACLIP, CaR, CLIPTrase, LaVG, SC-CLIP, ResCLIP, CLIP-DIY}. 
The next advancement, CLIPSurgery \cite{ClipSurgery}, sets up a parallel stream of processing blocks that take in the inputs of corresponding standard CLIP blocks, as well as the outputs of its direct predecessors, and combine them with value-similarity-based attention, followed by a residual connection. They experiment with implementing this adjacent path from varying depths and again note the benefits of calculating similarity in the same semantic space. The parallel blocks decidedly do not have multi-layer perceptrons (MLP-s) and their subsequent residual connections (\cref{fig:encoder_modified}) as Clip Surgery finds they introduce unnecessary noise into embeddings. 

NACLIP \cite{NACLIP} computes key-key similarity for attention and sums each individual patch's 2D-reordered attention weights with a 2D Gaussian kernel centered in that patch to encourage additional attention on its neighborhood. 
ClearCLIP \cite{ClearCLIP} modifies the last encoder layer by removing its MLP and both residual connections while using query-query self-self attention, where self-self denotes calculating similarity in the same semantic space, to obtain classifiable outputs.
GEM \cite{GEM} uses successive self-self attention modules to refine embeddings. The self-self attention blocks take the outputs of their self-self predecessor and its analogous standard transformer block to create a parallel stream of processing for producing the final embeddings. The normal embeddings are projected to query, key and values space respectively, and used in three parallel 2-stage self-self attention blocks. For brevity, we will go over only one of them. Query-projected embeddings represent all three inputs in the first stage self-self attention operation and serve as the query and key in the next stage self-self attention module, while the previously value-projected embeddings serve as values. The other blocks are analogous: replace query with key or value projected embeddings from the start. The three parallel outputs are average ensembled and summed with the previous block's outputs to produce final embeddings for the block.
SCLIP \cite{SCLIP} adopts the idea of using embeddings in the same semantic space for attention as well, this time calling it correlative self-attention, but only in the last encoder layer. They make the attention weights a sum of weights calculated from the similarity of query projected embeddings to themselves and the same for key projected embeddings respectively. 
Lastly, TagCLIP takes initial classification scores, obtained from a ViT modified for dense prediction (see first paragraph of section \ref{subsec:clip_based}), and then refines them further through attention; mixing results from similarly scored pixels. First, by using an attention mask to suppress interaction with non-similar patches or patches with low overall class confidence. The second step leverages the intermediary segmentation map from the last step to make class masks, utilizes them to mask over the original image, and classifies that with CLIP normally. The global per-region results are mixed with corresponding pixel scores.

\subsubsection{Leveraging Intermediate Transformer Layers} \label{subsubsec:clip_intermed}

Intermediate attention weights have proven to be advantageous as well. ITACLIP \cite{ITACLIP}, besides removing the MLP and corresponding residual connection in the last layer, combines SCLIP's final layer attention weights \cite{SCLIP} with attention weights extracted from intermediate layers. Moreover, they ensemble classification scores obtained on different geometrical and pixel-intensity transformations of the same image to increase performance. 
SC-CLIP \cite{SC-CLIP} first interpolates over "distracting" encodings in the penultimate layer by singling out patches with high Local Outlier Factor (LOF) \citep{LOF}, deeming they draw too much global attention during usual inter-token mixing. LOF is an anomaly detection method that measures the deviation of the local density of data points relative to its neighbors. An attention map from a middle layer is used for one final attention pass-through to obtain comparable embeddings.
ResCLIP \cite{ResCLIP} calculates temporary final layer attention weights as a combination of intermediate attention weights and self-self attention weights. In the following round of refinement, the corresponding segmentation map is used to constrain standard similarity calculation to patches sharing the same class. Those that do not are instead subject to a distance-based decay function on their similarity. Such scores are ensembled with the previous step's similarities to obtain attention weights and those again mixed with their predecessor for one final attention pass for outputs.

\subsubsection{Leveraging External Masks} \label{subsubsec:clip_mask}

Algorithmic ways to obtain region masks mark another sub-avenue of CLIP-based approaches. CaR \cite{CaR} consists of multiple rounds of pruning for classes which are actually present in the image. Grad-CAM \cite{gradCAM} constructs masks by back-propagating gradients for cosine similarity between each class and patch embedding. Mask-pooling obtains mask-wide embeddings suitable for matching to text such that classes not similar enough to any mask embedding are discarded each round until the class set remains unchanged.
CLIPTrase \cite{CLIPTrase} and LaVG \cite{LaVG}, on the other hand, leverage clustering to obtain masks. ClipTrase uses a sum of attention weights, quantifying q-q, k-k and v-v similarity respectively, in the last attention layer. Afterward, patches are clustered together using DBSCAN \cite{DBSCAN} based on their attention weights' similarity. DBSCAN groups closely packed points, and marks those in low-density regions as outliers; closeness is quantified by Euclidean distance. Mask-pooling over class-patch logits yields segmentation. 
LaVG uses normalcut where key embeddings of the last encoding layer correspond to nodes and their inter-cosine-similarity to weights. They mask-pool over SCLIP \cite{SCLIP} obtained image features to construct object prototypes ready for comparison and subsequent segmentation.

\subsubsection{Other Methods} \label{subsubsec:clip_misc}
Finally, we have ReCO \cite{ReCO} and DIY-CLIP \cite{CLIP-DIY} which uniquely differ from all other idea strains.
ReCO relies on a large, unlabeled, and robust dataset to be processed before use. The dataset serves to produce reference visual class embeddings with which target image patch embeddings can be compared to. The main drawback is that classes which have no image representation in the set-up dataset will not be recognized after deployment either.
Meanwhile, DIY-CLIP utilizes multi-scale partitionings of the original image. The image is divided into 1x1, 2x2, 3x3, etc. square "patches" subsequently fed to an unmodified CLIP encoder to produce per "patch" comparable visual embeddings. The similarity scores to class texts are spatially aggregated together if belonging to the same partitioning and upsampled so that all partitionings' scores are of the same resolution. FOUND \cite{FOUND} is leveraged to obtain foreground-background segmentation and used in conjunction with per-class multi-scale similarities to obtain final predictions.

\begin{figure}[htb]
    \centering
    \includegraphics[width=0.8\columnwidth]{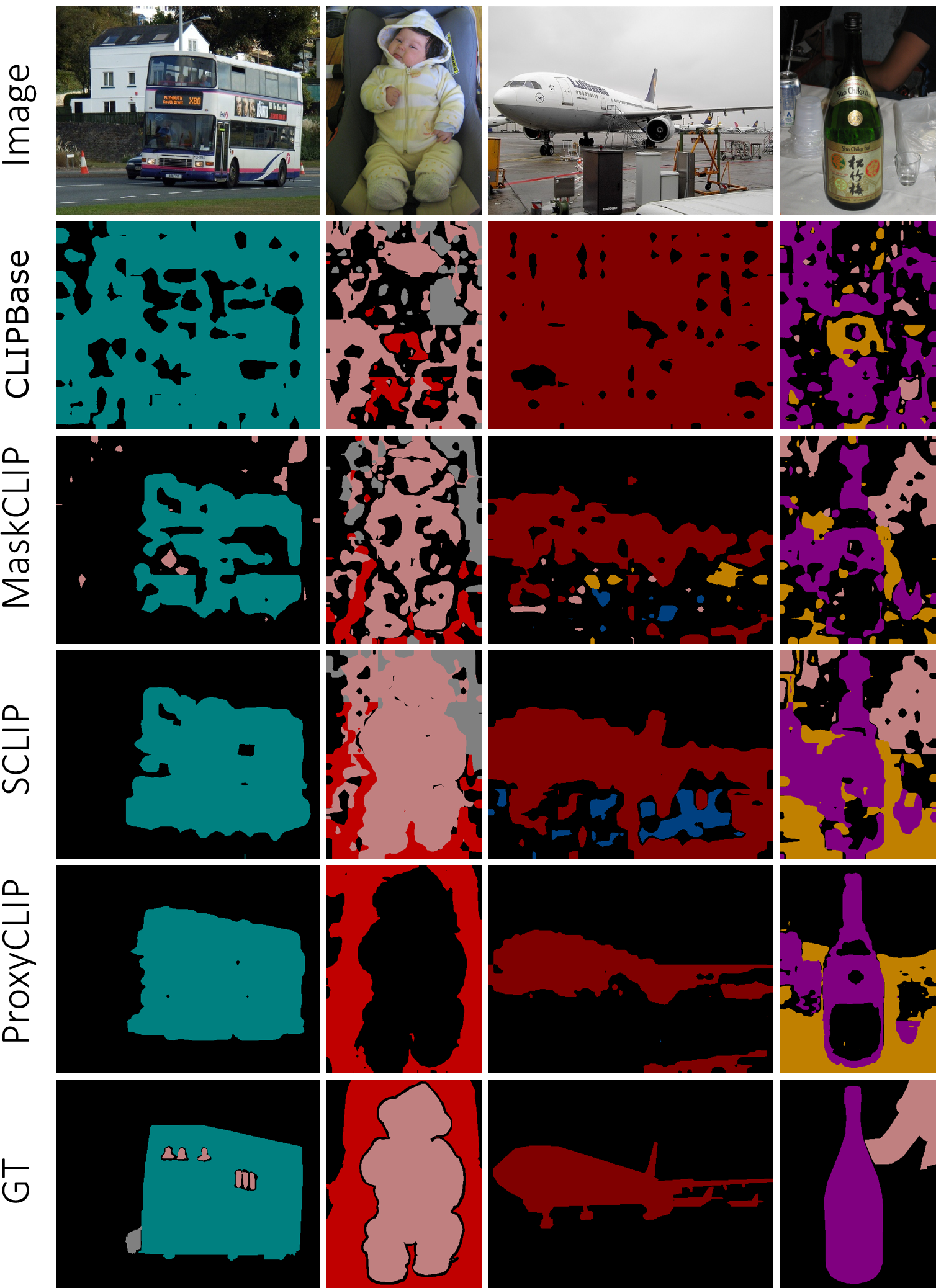}
    \caption{Qualitative comparison on the Pascal VOC21 \cite{voc} dataset of some standout methods \cite{MaskCLIP, SCLIP, ProxyCLIP}. GT denotes ground truth.}
    \label{fig:voc21s}
\end{figure}

\subsection{Improving CLIP with Auxiliary Visual Foundation Models} \label{subsec:vfms}

In contrast, \cite{ProxyCLIP, CASS, DBA-CLIP,Trident, CorrCLIP, TAG} rely on auxiliary VFM-s to exploit CLIP's visual embeddings, leveraging the fact that those VFM-s are, unlike CLIP, trained to encode spatial relationships as well. Two distinct strains of research have emerged: using external VFM-s to improve attention or to generate masks for mask-pooling. The models, besides TAG \citep{TAG} and DBA-CLIP \citep{DBA_CLIP}, follow the now-established convention of forgoing the last layer's MLP and both residual connections to remove the unnecessary noise they introduce into predictions.

\subsubsection{Improving Inter-Token Mixing} \label{subsubsec:vfm_attn}

ProxyCLIP \cite{ProxyCLIP} utilizes an auxiliary VMF's embeddings and performs attention weight calculation by quantifying their unmodified self-self similarity on account of their expected better spatial consistency. If the resolutions do not match, bilinear interpolation, respectful of 2D patch relationships, is used before the attention weights replace the originals in CLIP's last encoder layer and intermix normal CLIP value tokens. Although DINO ultimately performed best \citep{DINO}, during evaluation the authors tested out many more VFM-s \cite{MAE, SAM, SD, DINOv2} and found consistent improvement. In \cref{fig:voc21s} we can see the noticeable jump in region cohesion when advancing from purely CLIP-reliant methods to ProxyCLIP which drives home the usefulness of auxiliary VFM's spatial knowledge.

CASS \cite{CASS}  mixes attention weights extracted from the last layer of both CLIP and an auxiliary DINO model, trying to best match the individual attention heads such that they look at complimentary features and therefore instill object-level VFM knowledge into CLIP. Attention weight maps of individual heads are compared to weight maps of heads from the auxiliary VFM and matched so that wherever a CLIP attention head "did not look", the VFM head did, therefore enhancing the attended regions of patches.

\subsubsection{Leveraging External Localization Masks}
\label{subsubsec:vfm_maskpooling}

Mask-based VFM approaches first appeared in TAG \cite{TAG} which k-means clusters DINO \cite{DINO} image features into masks for pooling over GEM \citep{GEM} derived embeddings. The pixels inside are classified in accordance with mask-text embedding similarity.  
DBA-CLIP \cite{DBA-CLIP} uses masks generated by SAM \cite{SAM} to pool over MaskCLIP \cite{MaskCLIP} embeddings where some tokens are valued as more discriminative. These special tokens are singled out by looking at the average similarity between all tokens in a mask and reweighing MaskCLIP embeddings such that those with higher overall similarity are more pronounced. 
Trident \cite{Trident} first splits an image into overlapping sub-images, obtains their CLIP patch-level embeddings (see \ref{subsec:clip_based}) and inter-mixes them with corresponding attention weights of the last layer of a VFM. Taking the attention weights directly out of an auxiliary VFM marks a step above using the similarity of VFM embeddings themselves (like in ProxyCLIP \citep{ProxyCLIP}). The embeddings of individual windows are spliced together to obtain features for the whole image. Last-layer SAM attention weights from the original, untouched image are applied to convert features into classifiable per-patch embeddings.
CorrCLIP \cite{CorrCLIP} uses SAM to instead generate masks. Those with similar embeddings, constructed by pooling the sum of query and key DINO features of the same image, are combined. The masks are used on self-self attention weights calculated from the same, unmodified DINO features to constrain similarity between patches to only those that share a mask. Such weights apply a final attention operation on CLIP patch embeddings.

\subsection{Approaches Leveraging Generative Methods} \label{subsec:gen_methods}

The final dimension of research uses generative methods; either to obtain cross-attention maps between visual and textual information or prototypes for classes \cite{OvDiff, DiffSegmenter, FOSSIL, EmerDIFF, FreeSegDiff, FreeDA, CLIPer, MaskDiffusion}.  

OVDiff \cite{OvDiff} applies a foreground-background segmenter on generated class-example images. Pooling over their CLIP outputs yields representative features. Rather than comparing target image patches' similarity to text embeddings, the latter are used for retrieving corresponding prototype visual representations which facilitate segmentation.

DiffSegmentor \cite{DiffSegmenter} extracts cross and self-attention maps from an off-the-shelf pre-trained conditional latent diffusion model to construct masks: the former to produce initial class-aware masks, and the latter to better aggregate pixels belonging to the same category. 

EMERDIFF \cite{EmerDIFF} uses Stable Diffusion \cite{Stable_Diffusion} to obtain /32 semantically relevant embeddings which it then k-means clusters into low-res, class agnostic maps. Since the maps are the same resolution as the latent representation, they pick a particular mask and modify the corresponding latent embeddings by adding or subtracting small values to them, before letting that latent representation through the rest of the U-Net. By noticing which pixels are different between so-generated two images, they group them together. Since this is class-agnostic, they mask-pool over MaskCLIP \cite{MaskCLIP} features to recover the mask semantics. 

MaskDiffusion \cite{MaskDiffusion} also uses a frozen, pre-trained Stable Diffusion model. The underlying UNet is given latent image features obtained from a Variational Auto Encoder and CLIP-embedded text prompts. The outputs of all UNet layers are concatenated and pooled over by class-specific cross-attention maps, extracted from the UNet as well, to obtain respective class prototypes which are then compared to internal image features as in earlier approaches.

FreeSeg-Diff \cite{FreeSegDiff} extracts features from a diffusion model, clusters them, and uses the clusters to recover segments of potential objects. The original image is masked so that pixels not in a particular mask are zeroed out. Such images are passed through the normal CLIP visual tower and compared to text embeddings for final mask classification and subsequent segmentation. 

RIM \cite{RIM} uses Stable Diffusion to generate prototype images for each class, exploits SAM for masks and pools over DINOv2 \cite{DINOv2} features to make category reference features. To ensure the prototypes actually represent the wanted class, ie. the foreground, cross-attention maps for the specific classes are taken from the UNet, thresholded, and then the surviving points from such a binarized mask are used as prompts to SAM. During inference, SAM generates masks for the target picture, corresponding DINOv2 features get pooled and are then compared to reference features for final segmentation. 

FOSSIL \cite{FOSSIL} also uses generated textual and visual prototypes, where the former are used for retrieving relevant visual prototypes for a class, and the latter for classifying region embeddings. During setup, images are generated using captions passed to Stable Diffusion where, for each word, cross-attention is extracted to make the relevant binary mask for pooling over visual features of an image encoder. The textual reference is made by putting caption words into many templates and averaging over those representations along with the embedding of the whole caption: to preserve some of the context the class was in. Regions needed for inference are made by running normalcut to iteratively segment foreground objects based on visual embeddings. Features represent nodes, and cosine similarities the vertices. After each round, the graph is split into two disjoint parts after which the nodes belonging to the more prominent group are separated out and removed from consideration. Region-pooled representations are segmented like in previously discussed approaches.

FreeDA \cite{FreeDA} uses visual prototypes as well. During the set-up phase, a large number of captions are used to generate images with Stable Diffusion. Their DINO features get region pooled by nouns from the caption using relevant cross-attention maps. During inference, prototypes are retrieved by similarity to the target image embeddings. Image CLIP patch embeddings are compared to class embeddings, and DINO features are pooled by superpixel regions (constructed based on similar characteristics such as color, texture, or spatial proximity), then compared to visual prototypes. The two similarity results are combined to obtain the final segmentation. This bears a resemblance to FOSSIL, but, crucially, FreeDA does both retrieval and segmentation based on visual embeddings and uses superpixel regions.

Lastly, CLIPer obtains all intermediate layer attention maps and features, averages the attention maps, and replaces the last layer's attention weights with that. All the extracted embeddings are fed to that last layer, with the MLP and residual connections removed. By measuring similarity with those multiple embeddings and text encodings, CLIPer generates an averaged similarity map: coarse and patch-level. To refine the predictions, they input the original image to a Stable Diffusion model with an empty prompt and extract the corresponding multi-head attention maps at the highest spatial resolution. The attention maps are fused by matrix chain multiplication across the attention heads and the fusion serves to obtain the final segmentation result by remixing the coarse segmentation scores.

\begin{table*}[!ht]
    \centering
    \footnotesize
    \begin{tabular}{l|cccccccc}
       \toprule
        & VOC21 & Context60 & COCO Object & VOC20 & CityScapes & Context59 & ADE20K & COCO Stuff \\
        \hline
        \multicolumn{1}{l|}{\textit{Purely CLIP based}}  & & & & & & & & \\  \hline
        \multicolumn{1}{c|}{\textit{Refine inter-token mixing}} & ~ & ~ & ~ & ~ & ~ & ~ & ~ & ~ \\
        CLIPBase \textsuperscript{\cite{SCLIP}} & 18.8 & 9.9 & 8.1 & 49.4 & 6.5 & 11.1 & 3.1 & 5.7  \\
        MaskCLIP\textsuperscript{\cite{SCLIP}}  & 43.4 & 23.2 & 20.6 & 74.9 & 24.9 & 26.4 & 11.9 & 16.7 \\  
        CLIP Surgery & - & - & - & - & 31.4 & 29.3 & - & 21.9 \\  
        NACLIP & 58.9 & 32.2 & 33.2 & 79.7 & 35.5 & 35.2 & 17.4 & 23.3 \\  
        ClearCLIP & 51.8 & 32.6 & 33.0 & 80.9 & 30.0 & 35.9 & 16.7 & 23.9 \\  
        GEM  & 46.2 & - & - & - & - & 32.6 & 15.7 & - \\  
        SCLIP & 59.1 & 30.4 & 30.5 & 80.4 & 32.2 & 34.2 & 16.1 & 22.4 \\  
        TagCLIP\textsuperscript{\cite{ITACLIP}} & 64.8 & - & 33.5 & - & - & - & - & 18.7 \\

        \hline
        \multicolumn{1}{c|}{\textit{Leverage intermediate layer outputs}} & ~ & ~ & ~ & ~ & ~ & ~ & ~ & ~ \\
        ITACLIP  & 65.6 & 36.0 & 36.4 & - & 39.2 & - & - & 26.3 \\  
        SC-CLIP & 64.6 & 36.8 & 37.7 & 84.3 & 41.0 & 40.1 & 20.1 & 26.6 \\  
        ResCLIP$_{\text{on top of NACLIP}}$  & 61.1 & 33.5 & 35.0 & 86.0 & 35.9 & 36.8 & 18.0 & 24.7 \\  

        \hline
        \multicolumn{1}{c|}{\textit{Leverage non-ML obtained masks}} & ~ & ~ & ~ & ~ & ~ & ~ & ~ & ~ \\
        CaR\textsubscript{L/14*} & 67.6 & 30.5 & 36.6 & 91.4 & - & 39.5 & 17.7 & - \\  
        CLIPtrase & 53.0 & 30.8 & 44.8 & 81.2 & - & 34.9 & 17.0 & 24.1 \\  
        LaVG & 62.1 & 31.6 & 34.2 & 82.5 & - & 34.7 & 15.8 & 23.2 \\

        \hline
        \multicolumn{1}{c|}{\textit{Other approaches}} & ~ & ~ & ~ & ~ & ~ & ~ & ~ & ~ \\
        ReCO\textsuperscript{\cite{SCLIP}}\textsubscript{L/14}  & 25.1 & 19.9 & 15.7 & 57.7 & 21.6 & 22.3 & 11.2 & 14.8 \\  
  
        DIY-CLIP &  59.0 & - & 30.4 & - & - & - & - & - \\ 

        \hline
        \multicolumn{1}{l|}{\textit{Use VFM-s alongside CLIP}}  & & & & & & & & \\  \hline
        \multicolumn{1}{c|}{\textit{Refine inter-token mixing with VFM-s}} & ~ & ~ & ~ & ~ & ~ & ~ & ~ & ~ \\
        
        ProxyCLIP  & 61.3 & 35.3 & 37.5 & 80.3 & 38.1 & 39.1 & 20.2 & 26.5 \\ 
        CASS & 65.8 & 36.7 & 37.8 & 87.8 & 39.4 & 40.2 & 20.4 & 26.7 \\

        \hline
        \multicolumn{1}{c|}{\textit{Leverage VFM-s for maskpooling}} & ~ & ~ & ~ & ~ & ~ & ~ & ~ & ~ \\
        TAG\textsubscript{L/14} & - & - & - & 56.9 & - & 20.2 & 6.6 & - \\ 
        DBA-CLIP  & 74.3 & - & 43.8 & 86.7 & 44.1 & - & - & 29.3 \\ 
        Trident  & 67.1 & 38.6 & 41.1 & 84.5 & 42.9 & 42.2 & 21.9 & 28.3 \\ 
        CorrCLIP  & 72.5 & 42.0 & 43.7 & 88.7 & 48.3 & 46.2 & 25.3 & 30.6 \\ 

        \hline
        \multicolumn{1}{l|}{\textit{Use generative methods alongside CLIP}}  & & & & & & & & \\  \hline

        OVDiff & - & - & 34.6 & 66.3 & - & 29.7 & - & - \\
        DiffSegmenter & 60.1 & 27.5 & 37.9 & - & - & - & - & - \\ 
        EMERDIFF$_{\text{on top of MaskCLIP}}$ & - & - & - & - & 26.5 & - & 15.9 & - \\ 
        MaskDiffusion & - & - & - & - & 28.5 & - & - & - \\ 
        FreeSeg-Diff & 53.3 & - & 31.0 & - & - & - & - & - \\ 
        RIM & - & - & 44.5 & 77.8 & - & 34.3 & - & - \\ 
        FOSSIL & - & - & - & - & 23.2 & 35.8 & 18.8 & 24.8 \\ 
        FreeDA & - & - & - & 85.6 & 36.7 & 43.1 & 22.4 & 27.8 \\ 
        CLIPer  & 65.9 & 37.6 & 39.0 & 85.2 & - & 41.7 & 21.4 & 27.5 \\ 
        
        \bottomrule
    \end{tabular}
    \caption{Open-vocabulary semantic segmentation mIoU comparison of training-free methods on eight benchmarks: PASCAL VOC21/20 \cite{voc}, PASCAL CONTEXT60/59\cite{context}, COCO Object/Stuff \citep{cocostuffandobject}, Cityscapes \cite{cityscapes} and ADE20K \cite{ade20k}. VOC20, Context59, and COCO Stuff represent the original datasets with the "background" class removed. All reported results are without post-processing methods (e.g. PAMR \cite{PAMR}) and, except if denoted with a subscript that specifies the ViT model, use a ViT-B/16 CLIP to ensure a fairer comparison. * CaR uses a ViT-L/14 CLIP instance \textit{in addition} to the ViT-B/16. Unless indicated with a superscript, the results were imported from the original papers. The baseline CLIP, MaskCLIP, and ReCO evaluations were instead taken from \cite{SCLIP} since the original papers only included results on 2 and 3 benchmarks, respectively. TagCLIP results are from the reproduction in \cite{ITACLIP} as the original did not include comparable results.}

    \label{tab:divided_tfovss}
\end{table*}

\begin{table*}[!ht]
    \centering
    \footnotesize
    \begin{tabular}{l|cccccccc}
       \toprule
        & Voc21 & Context60 & COCO Object & Voc20 & CityScapes & Context59 & ADE20K & COCO Stuff \\
        \toprule
        \multicolumn{1}{l|}{\textit{Purely CLIP based}}  & & & & & & & & \\  \hline
        \textcolor{gray}{NACLIP - ViT-B} & \textcolor{gray}{58.9} & \textcolor{gray}{32.2} & \textcolor{gray}{33.2} & \textcolor{gray}{79.7} & \textcolor{gray}{35.5} & \textcolor{gray}{35.2} & \textcolor{gray}{17.4} & \textcolor{gray}{23.3} \\ 
        NACLIP - PAMR & 57.9 & - & - & - & - & 36.4 & - & - \vspace{1.5pt}\\ 
        
        \textcolor{gray}{ClearCLIP - ViT-B} & \textcolor{gray}{51.8} & \textcolor{gray}{32.6} & \textcolor{gray}{33.0} & \textcolor{gray}{80.9} & \textcolor{gray}{30.0} & \textcolor{gray}{35.9} & \textcolor{gray}{16.7} & \textcolor{gray}{23.9} \\ 
        ClearCLIP & - & - & - & 80.0 & 27.9 & 29.6 & 15.0 & 19.9 \vspace{1.5pt}\\ 
        \textcolor{gray}{GEM - CLIP - ViT-B} & \textcolor{gray}{46.2} & \textcolor{gray}- & \textcolor{gray}- & \textcolor{gray}- & \textcolor{gray}- & \textcolor{gray}{32.6} & \textcolor{gray}{15.7} & \textcolor{gray}- \\ 
        GEM - CLIP & 44.6 & - & - & - & - & 28.6 & - & - \vspace{1.5pt}\\ 
        \textcolor{gray}{ITACLIP - ViT-B} & \textcolor{gray}{65.6} & \textcolor{gray}{36.0} & \textcolor{gray}{36.4} & \textcolor{gray}- & \textcolor{gray}{39.2} & \textcolor{gray}- & \textcolor{gray}- & \textcolor{gray}{26.3} \\ 
        ITACLIP - PAMR  & 53.3 & - & - & - & - & - & - & - \vspace{1.5pt}\\ 
        \textcolor{gray}{SC-CLIP - ViT-B} & \textcolor{gray}{64.6} & \textcolor{gray}{36.8} & \textcolor{gray}{37.7} & \textcolor{gray}{84.3} & \textcolor{gray}{41.0} & \textcolor{gray}{40.1} & \textcolor{gray}{20.1} & \textcolor{gray}{26.6} \\ 
        SC-CLIP & 65.0 & 36.9 & 40.5 & 88.3 & 41.3 & 40.6 & 21.7 & 26.9 \vspace{1.5pt}\\ 
        \textcolor{gray}{ResCLIP - ViT-B} & \textcolor{gray}{61.1} & \textcolor{gray}{33.5} & \textcolor{gray}{35.0} & \textcolor{gray}{86.0} & \textcolor{gray}{35.9} & \textcolor{gray}{36.8} & \textcolor{gray}{18.0} & \textcolor{gray}{24.7} \\
        ResCLIP & 54.1 & 30.9 & 32.5 & 85.5 & 33.7 & 34.5 & 18.2 & 23.4 \vspace{1.5pt}\\

        \hline
        \multicolumn{1}{l|}{\textit{Use VFM-s alongside CLIP}}  & & & & & & & & \\  \hline
        
        \textcolor{gray}{ProxyCLIP - ViT-B} & \textcolor{gray}{61.3} & \textcolor{gray}{35.3} & \textcolor{gray}{37.5} & \textcolor{gray}{80.3} & \textcolor{gray}{38.1} & \textcolor{gray}{39.1} & \textcolor{gray}{20.2} & \textcolor{gray}{26.5} \\ 
        ProxyCLIP  & 60.6 & 34.5 & 39.2 & 83.2 & 40.1 & 37.7 & 22.6 & 25.6 \vspace{1.5pt}\\ 
        \textcolor{gray}{CASS - ViT-B} & \textcolor{gray}{65.8} & \textcolor{gray}{36.7} & \textcolor{gray}{37.8} & \textcolor{gray}{87.8} & \textcolor{gray}{39.4} & \textcolor{gray}{40.2} & \textcolor{gray}{20.4} & \textcolor{gray}{26.7} \\ 
        CASS & 62.1 & - & - & - & - & 39.1 & - & 26.3 \vspace{1.5pt}\\ 
        \textcolor{gray}{Trident - ViT-B} & \textcolor{gray}{67.1} & \textcolor{gray}{38.6} & \textcolor{gray}{41.1} & \textcolor{gray}{84.5} & \textcolor{gray}{42.9} & \textcolor{gray}{42.2} & \textcolor{gray}{21.9} & \textcolor{gray}{28.3} \\ 
        Trident & 70.8 & 40.1 & 42.2 & 88.7 & 47.6 & 44.3 & 26.7 & 28.6 \vspace{1.5pt}\\ 
        \textcolor{gray}{CorrCLIP - ViT-B} & \textcolor{gray}{72.5} & \textcolor{gray}{42.0} & \textcolor{gray}{43.7} & \textcolor{gray}{88.7} & \textcolor{gray}{48.3} & \textcolor{gray}{46.2} & \textcolor{gray}{25.3} & \textcolor{gray}{30.6} \\ 
        CorrCLIP & 73.2 & 41.0 & 46.0 & 90.6 & 49.0 & 47.5 & 29.1 & 32.0 \vspace{1.5pt}\\ 

        \hline
        \multicolumn{1}{l|}{\textit{Use generative methods alongside CLIP}}  & & & & & & & & \\  \hline

        \textcolor{gray}{FreeDA - ViT-B} & \textcolor{gray}- & \textcolor{gray}- & \textcolor{gray}- & \textcolor{gray}{85.6} & \textcolor{gray}{36.7} & \textcolor{gray}{43.1} & \textcolor{gray}{22.4} & \textcolor{gray}{27.8} \\ 
        FreeDA & - & - & - & 87.9 & 36.7 & 43.5 & 23.2 & 28.8 \vspace{1.5pt}\\ 
        \textcolor{gray}{CLIPer - ViT-B} & \textcolor{gray}{65.9} & \textcolor{gray}{37.6} & \textcolor{gray}{39.0} & \textcolor{gray}{85.2} & \textcolor{gray}- & \textcolor{gray}{41.7} & \textcolor{gray}{21.4} & \textcolor{gray}{27.5} \\ 
        CLIPer  & 69.8 & 38.0 & 43.3 & 90.0 & - & 43.6 & 24.4 & 28.7 \vspace{1.5pt}\\ 
        
        \bottomrule
    \end{tabular}
    \caption{Open-vocabulary mIoU comparison of training-free methods with respect to backbone size on eight benchmarks: PASCAL VOC21/20 \cite{voc}, PASCAL CONTEXT60/59\cite{context}, COCO Object/Stuff \citep{cocostuffandobject}, Cityscapes \cite{cityscapes} and ADE20K \cite{ade20k}. VOC20, Context59 and COCO Stuff represent original the datasets with the "background" class removed. The grayed-out rows represent results from the original papers with the ViT-B/16 CLIP backbone and were taken from \cref{tab:divided_tfovss}, while the full-color results are with the ViT-L/14 (or ViT-H/14 in the isolated case of Trident). Unless stated after the dash, the results do not utilize any additional post-processing.}
    \label{tab:vitl_ovss}
\end{table*}

\section{Current Problems and Future Direction}

This section will discuss some less researched and potentially promising directions in training-free open-vocabulary semantic segmentation.

\subsection{Full Potential of ViT CLIP}
According to \cref{tab:divided_tfovss}, the current leaders in performance are DBA-CLIP \cite{DBA_CLIP} and CorrCLIP \cite{CorrCLIP}, both of which rely on SAM \cite{SAM} or DINO \cite{DINO} to mask-pool over CLIP \cite{CLIP} features. However, purely CLIP-reliant approaches, eg. SC-CLIP \cite{SC-CLIP} or CaR \cite{CaR}, are not far off which raises the question of CLIP's maximum potential. Given the close performances, as well as probable inference time savings when not using additional models, there is still more to be found.

\subsection{Background Segmentation Relies on Context}
The examined methods all have slightly different takes on handling the background category present in some datasets. Some build its textual description from common things found in the background (eg. sky, wall, tree, wood, grass, road...), which may prove problematic when a dataset has both "sky" and "background" classes, while some use vision-based methods to distinguish between background and foreground pixels, which may prevent them from segmenting things commonly found in the background.

All of these rely on either the researchers' understanding of what the background is, or how some other model has come to understand it. This is compounded upon by the datasets, as they do not unanimously agree, but we will stick to specifically talking about methods underexploring the ability of their models to refuse classification of pixels. In essence, to lump them together as \emph{non-relevant}, fundamentally \emph{background} to the class set, without relying on prior knowledge of the background. To instead base segmentation purely on the other, \emph{non-background}, class prompts.

\subsection{Prevalent Reliance on Window Slide Inference}
During inference, images are most commonly divided into overlapping windows the size the underlying ViT was trained on; stemming from transformers' sensitivity to resolution. This, in turn, incurs additional time both for the individual inferences and the subsequent aggregation. During our research, we noticed that inference over an image in one pass possibly has unexplored potential and a lot of room for improvement.

\subsection{Worse Results With Larger Backbones for Some Approaches}
As evidenced by table \ref{tab:vitl_ovss}, a non-insignificant number of methods suffer degraded performance when using larger backbone models: ClearCLIP \cite{ClearCLIP}, GEM \cite{GEM}, ITACLIP \cite{ITACLIP}, ResCLIP \cite{ResCLIP} and CASS \cite{CASS}. Interestingly, the first four belong to the category of purely CLIP-based method (while CASS uses VFM-s), which might suggest a limitation of how CLIP utilizes the embedding space in larger models. If global image information can be gleaned from a smaller share of all tokens or their overall content, the rest (be it tokens or content) will not be forced to carry useful substance. Other methods, which rely on auxiliary models to exploit CLIP embeddings, do not suffer the same problem.

\subsection{Convolutional CLIP}
While transformer-based CLIP is no doubt powerful, there is an almost equally successful convolutional version based on the ConvNext \citep{ConvNext}  architecture, and yet, to our knowledge, no work has tried adapting it for semantic segmentation. These models might prove to be more suited to encoding per-pixel embeddings that carry local information due to their nature. Whether by developing methods based solely on ConvNext CLIP, or employing it in tandem with ViT, we believe there is untapped potential for study.

\section{Conclusion}

In this survey, we provide an extensive review and careful explanation of both the most current methods and the approaches that allowed training-free semantic segmentation to reach them. We touch upon task-important model archetypes, define the examined problem, and present the overarching strains of guiding ideas in this area of research. Furthermore, we look at over 30 training-free methods to provide a deeper understanding and spotlight their idiosyncrasies and advancements, as well as compile an exhaustive results table for easy performance comparison. Lastly, we turn our attention to possibly underexplored avenues and ruminate over potentially fruitful directions for future work. We hope for this survey to be a useful introductory piece that allows researchers to quickly get up to speed in the area and sparks even more interest.

{
    \small
    \bibliographystyle{ieeenat_fullname}
    \bibliography{main}
}

\end{document}